\documentclass{aamas2016}

\usepackage{algorithm,algpseudocode}

\usepackage{graphicx}
\usepackage{caption}
\usepackage{subcaption}

\usepackage{subfig}

\makeatletter
\def\@copyrightspace{\relax}
\makeatother

\begin{document}


\title{Feature Selection as a Multiagent Coordination Problem}



%
%
%
%

%

\numberofauthors{4}

\author{
%
\alignauthor
Kleanthis Malialis\\
       \affaddr{Dept. of Computer Science\\University College London, UK}\\
       \email{k.malialis@ucl.ac.uk}
\alignauthor
Jun Wang\\
       \affaddr{Dept. of Computer Science\\University College London, UK}\\
       \email{j.wang@ucl.ac.uk}
\and
\alignauthor
Gary Brooks\\
       \affaddr{Massive Analytic\\London, UK}\\
       \email{gary.brooks@\\massiveanalytic.com}
\alignauthor
George Frangou\\
       \affaddr{Massive Analytic\\London, UK}\\
       \email{george.frangou@\\massiveanalytic.com}
}

\maketitle

\begin{abstract}
Datasets with hundreds to tens of thousands features is the new norm. Feature selection constitutes a central problem in machine learning, where the aim is to derive a representative set of features from which to construct a classification (or prediction) model for a specific task. Our experimental study involves microarray gene expression datasets; these are high-dimensional and noisy datasets that contain genetic data typically used for distinguishing between benign or malicious tissues or classifying different types of cancer. In this paper, we formulate feature selection as a multiagent coordination problem and propose a novel feature selection method using multiagent reinforcement learning. The central idea of the proposed approach is to ``assign'' a reinforcement learning agent to each feature where each agent learns to control a single feature; we refer to this approach as MARL. Applying this to microarray datasets creates an enormous multiagent coordination problem between thousands of learning agents. To address the scalability challenge we apply a form of reward shaping called CLEAN rewards. We compare in total nine feature selection methods, including state-of-the-art methods, and show that the proposed method using CLEAN rewards can significantly scale-up, thus outperforming the rest of learning-based methods. We further show that a hybrid variant of MARL achieves the best overall performance.
\end{abstract}





\section{Introduction}
Datasets nowadays are becoming increasingly rich with hundreds to tens of thousands of features available. This explosion of information, however, contains features that are irrelevant (i.e. have little predictive or classification power), noisy and redundant. A central problem in machine learning is to derive a representative subset of these features from which to construct a classification (or prediction) model for a specific task \cite{hall1999}.

Feature selection offers many potential benefits \cite{guyon2003}. It can significantly boost classification performance (or prediction accuracy), create human-understandable results and provide insight to the data by returning only the top classifiers for a particular task. It can further facilitate data visualisation, reduce storage requirements and execution runtime of machine learning algorithms.

There are three types of feature selection methods. Filters \cite{guyon2003} use variable ranking techniques that rely on statistical measures to select top-ranked features. Filtering is performed as a pre-processing step i.e. independent of the classification algorithm. Wrappers \cite{chandrashekar2014} do take into consideration the algorithm's performance as the objective function in order to evaluate feature subsets. This is known to be NP-hard as enumeration of all possible feature subsets is intractable. A hybrid filter-wrapper approach combines the advantages of both worlds. Alternatively, in embedded \cite{guyon2003,chandrashekar2014} approaches, the feature selection process is built-in or ``embedded'' in the classifier directly.

Our experimental study involves microarray gene expression datasets. Microarray datasets are high-dimensional (thousands of features) datasets that contain genetic data typically used for distinguishing between benign or malicious tissues or classifying different types of cancer. The contributions made by this work are the following.

We formulate feature selection as a multiagent coordination problem and propose a novel wrapper feature selection method using multiagent reinforcement learning. The central idea of the proposed approach is to ``assign'' a learning agent to each feature. Each agent controls a single feature and learns whether it will be included in or excluded from the final feature subset; we refer to this approach as MARL.

Applying this to microarray datasets creates a large multiagent coordination problem between thousands of learning agents. To address the scalability challenge we apply CLEAN rewards \cite{holmesparker2014,colby2015}, a form of reward shaping, that removes the exploratory noise caused by other learning agents and has been shown to achieve excellent empirical results in a variety of domains.

We compare in total nine feature selection methods including state-of-the-art methods. Specifically, a baseline, two filter, three wrapper and three hybrid methods were included in our comparison. We show that the proposed CLEAN method significantly outperforms the wrapper methods, which severely suffer from scalability issues. The CLEAN method is the only wrapper that, out of thousands of features, can learn a feature subset whose size is below a desired boundary. Furthermore, as expected, the proposed wrapper MARL method suffers from scalability issues, however, it is demonstrated that a hybrid variant of MARL achieves the best overall performance.

The organisation of the paper is as follows. Section~\ref{sec:preliminaries} describes the background material necessary to understand the contributions made by this work. Section~\ref{sec:related} presents the work that is related to ours. Section~\ref{sec:MARL} introduces and describes in detail our proposed approach. Learning and evaluation results are discussed in Section~\ref{sec:learning} and \ref{sec:eval} respectively. A conclusion is presented in Section~\ref{sec:conclusion}.

\section{Preliminaries}\label{sec:preliminaries}

\subsection{Reinforcement Learning}
Reinforcement learning is a paradigm in which an active decision-making agent interacts with its environment and learns from reinforcement, that is, a numeric feedback in the form of reward or punishment \cite{sutton1998}. The feedback received is used to improve the agent's actions. The concept of an iterative approach constitutes the backbone of the majority of reinforcement learning algorithms. In this work we are interested in stateless tasks, where reinforcement learning is used by the agents to estimate, based on previous experience, the expected reward associated with individual or joint actions.

A popular reinforcement learning algorithm is Q-learning \cite{watkins1992}. In the stateless setting, the Q-value $Q(a)$ provides the estimated expected reward of performing (individual or joint) action $a$. An agent updates its estimate $Q(a)$ using:
\begin{equation}\label{eq:Qupdate}
  Q(a)\leftarrow Q(a)+\alpha[r - Q(a)]
\end{equation}
where $a$ is the action taken resulting in reward $r$, and $\alpha \in [0,1]$ is the rate of learning.

Applications of multiagent reinforcement learning typically take one of two approaches; multiple individual learners (ILs) or joint action learners (JALs) \cite{claus1998}. Multiple ILs assume any other agents to be part of the environment and so, as the others simultaneously learn, the environment appears to be dynamic as the probability of transition when taking action $a$ in state $s$ changes over time. To overcome the appearance of a dynamic environment, JALs were developed that extend their value function to consider for each state the value of each possible combination of actions by all agents. The consideration of the joint action causes an exponential increase in the number of values that must be calculated with each additional agent added to the system. Therefore, as we are interested in scalability, this work focuses on multiple ILs and not JALs.

The exploration-exploitation trade-off constitutes a critical issue in the design of a reinforcement learning agent. It aims to offer a balance between the exploitation of the agent's knowledge and the exploration through which the agent's knowledge is enriched. A common method of doing so is $\epsilon$-greedy \cite{sutton1998}, where the agent behaves greedily most of the time, but with a probability $\epsilon$ it selects an action randomly. To get the best of both exploration and exploitation, it is advised to reduce $\epsilon$ over time \cite{sutton1998}.

The typical approach in MARL is to provide agents with a global reward. This reward is in fact the system performance used as a learning signal, thus allowing each agent to act in the system's interest. The global reward, however, includes a substantial amount of noise due to exploratory actions \cite{holmesparker2014}. This is caused by the fact that learning agents are unable to distinguish what parts of the global reward signal are caused by true environmental dynamics, and what parts are caused by exploratory action noise caused by other agents. CLEAN rewards \cite{holmesparker2014,colby2015} were designed to remove exploratory action noise and are described in the next section.

\subsection{CLEAN Rewards}\label{sec:preliminaries_CLEAN}
CLEAN rewards \cite{holmesparker2014,colby2015} were introduced to remove exploratory noise present in the global reward. This is achieved because the exploration for each agent is performed offline i.e. privately. Specifically, at each learning episode, each agent executes an action by following its greedy policy (i.e. without exploration); then all the agents receive a global reward. Each agent then privately computes the global reward it would have received had it executed an exploratory action, while the rest of the agents followed their greedy policies.

CLEAN rewards were defined as follows \cite{holmesparker2014}:
\begin{equation}\label{eq:CLEAN}
 C_i = G(a - a_i + c_i) - G(a)
\end{equation}
where $a$ is the joint action executed when all agents followed their greedy policies, $a_i$ is the greedy action executed by agent $i$, $c_i$ is the counterfactual (offline) action taken by agent $i$ following $\epsilon$-greedy, $G(a)$ is the global reward received when all agents executed their greedy policies and $G(a - a_i + c_i)$ is the counterfactual (offline) reward agent $i$ would have received, had it executed the counterfactual action $c_i$, instead of action $a_i$, while the rest of the agents followed their greedy policies. Each agent then uses the following formula to update its Q-values:
\begin{equation}\label{eq:CLEANupdate}
  Q(c_i)\leftarrow Q(c_i)+\alpha[C_i - Q(c_i)]
\end{equation}

In this manner, CLEAN rewards remove the exploratory noise caused by other agents and allow each agent to effectively determine which actions are beneficial or not. CLEAN rewards have achieved superior empirical results in a variety of domains such as in UAV communication networks  \cite{holmesparker2014}.

\subsection{Feature Selection}
Feature selection is defined as the problem of discarding information that is irrelevant (i.e. have little predictive or classification power), noisy and redundant, thus considering only a subset of the features. In other words, the goal of feature selection is to identify the best classifiers (or predictors) for a particular classification (or prediction) task.

Formally \cite{ng1998}, let $X$ be the fixed $f$-dimensional input space, where $f$ is the number of features. Let $T$ be the $m \times f$-dimensional space (subspace of $X$) that represents the training set $T = \{x^i,y^i\}_{i=1}^{m}$, where $m$ is the number of training examples. Each example in $T$ consists of a $f$-dimensional input vector $x^i = [x^i_1, x^i_2, ..., x^i_f]$ drawn i.i.d. from some fixed distribution $D_X$ over $X$, and an output class label $y^i$. We denote by $C$ the $m$-dimensional vector of the class labels $C = [y^1, y^2, ..., y^m]$. Without any loss of generality, in this work we focus on binary classification problems, and assume a fixed binary concept $g : X \longmapsto \{0,1\}$ such that $y^i = g(x^i) \in \{0,1\}$.

Note that $g$ uses all $f$ features, but it is hoped that it can be approximated well (in terms of the generalisation error) by a hypothesis function that depends only on a small subset of the $f$ features. Let F be the set of all $f$ features $F = \{F_1, F_2, ..., F_f \}$, and $S \subseteq F$ be a subset of it. We denote by $x^i |_S$ the input vector with all the features not in $S$ eliminated, and similarly,  let $X|_S$ be the input space with all the features not in $S$ eliminated.

The aim of feature selection is to find a feature subset $S$, such that a hypothesis $h : X|_S \longmapsto \{0,1\}$ which is defined only in the subspace of features $X|_S$, can be extended to $X$. For any hypothesis $h$ the generalisation error is defined as $\epsilon (h) = Pr_{x \in D_X} [  h(x) \neq g(x) ]$ and the empirical error on the training set $T$ is $\hat{\epsilon} (h) = \frac{1}{|T|} | \{ (x,y) \in T | h(x) \neq y \} |$.

\section{Related Work}\label{sec:related}
Traditionally, there are three categories of feature selection algorithms. \textbf{Filters} \cite{guyon2003} use variable ranking techniques that rely on statistical measures (such as Pearson's correlation coefficient and mutual information) to select top-ranked features. It is important to note that filtering is performed as a data pre-processing step i.e. independent of the choice of the classification algorithm. Filters have been demonstrated to perform well in many cases \cite{chandrashekar2014} and they are also computationally cost-effective and fast \cite{chandrashekar2014}.

Let $T|_F$ and $C$ be the training set with the class labels vector. When each feature $F_i \in F$ is considered individually we refer to it as a univariate filter. An example of this is the univariate correlation-based feature selection (uCFS) \cite{guyon2003} where the Pearson's correlation between each feature space $T|_{F_i \in F}$ and the target class $C$ is taken. Based on the square of this score the top-ranked features are selected; the square is considered so negatively correlated features are included.

When a subset of features is considered we refer to it as a multivariate filter. A popular example of this is the minimal-redundancy maximal-relevance (mRMR) method \cite{peng2005}. mRMR searches for a subset $S \subseteq F$ that maximises relevance by maximising the mutual information between each individual feature space $T|_{F_i \in S}$ and the class $C$, and at the same time minimises redundancy by minimising the mutual information between any two feature spaces $T|_{F_i \in S}$ and $T|_{F_j \in S}$ $(i \neq j)$ within the subset. Another example of this category is the multivariate CFS \cite{hall1999}.

\textbf{Wrappers} \cite{chandrashekar2014} consider the classification algorithm's performance (e.g. accuracy) as the objective function in order to evaluate feature subsets. This is known to be NP-hard as enumeration of all $2^f$ possible feature subsets is intractable; for this reason heuristic search algorithms are employed. Wrappers typically outperform filters because they take into account the feedback from the classifier, at the expense of being computationally intensive and slow. Examples include hill climbing algorithms \cite{hall1999} such as forward selection and backward elimination, best-first search \cite{hall1999}, sequential floating forward selection \cite{chandrashekar2014}, Monte Carlo tree search \cite{gaudel2010}, neural networks \cite{kabir2010} and genetic algorithms \cite{glaab2012,chandrashekar2014,soufan2015}.

A \textbf{hybrid} filter-wrapper approach, where a filter is first used to bring down the number of features and then a wrapper is applied, combines the advantages of both worlds. Alternatively, in \textbf{embedded} \cite{guyon2003,chandrashekar2014} approaches, the feature selection process is built-in or ``embedded'' in the classifier directly; popular examples are decision trees \cite{breiman1984} and regularization methods \cite{ng2004}.

\section{MARL Wrapper}\label{sec:MARL}
In this paper we formulate feature selection as a multiagent coordination problem and propose a novel wrapper feature selection method using multiagent reinforcement learning. The central idea of the proposed approach is to ``assign'' a learning agent to each feature. Thus, the total number of learning agents is equal to the number of features.

The purpose of feature selection is to discard irrelevant, noisy and redundant features, by only proposing a subset of the features. Each reinforcement learning agent is therefore allowed to control a single feature; in other words, to decide whether a feature will be included in or excluded from the final feature subset.

Each agent has two actions; these are, ``0'' and ``1''. Action ``0'' represents a feature that is ``disabled'' or ``off'' i.e. it is not included in the final feature subset. Analogously, action ``1'' represents a feature that is ``enabled'' or ``on'' i.e. it is included in the feature subset. Therefore, a joint action is a bitstring (i.e. a sequence of ``0''s and ``1''s) and in order to get the feature subset we need to extract only the ``1''s.

\begin{algorithm}[t]             
\caption{Feature Selection}   
\label{alg:kFoldCV}                  
\begin{algorithmic}[1]           

\Function{kFoldCV\_FS}{$D, L,\lambda,k,b$}
	\Statex Input parameters: 
	\Statex $D$: dataset, $L$: classification algorithm
	\Statex $\lambda$: split percentage, $k$ number of folds
	\Statex $b$: upper boundary for feature subset's size

	\State randomly split D into the training $T$ ($(1-\lambda)\%$) and evaluation $E$ ($\lambda\%$) sets
	\State randomly split $T$ into $k$ disjoint folds $\{T_1,...,T_k\}$ of $m/k$ training examples each
	
	\State get feature subset $S = FS\_MARL(\{T_1,...,T_k\})$\Comment Feature selection (Algorithm~\ref{alg:fs_clean})
	
	\State learn hypothesis $h = L(T|_S)$\Comment Training
	\State get performance $P$ of $h$ on $E|_S$\Comment Evaluation on unseen dataset
	\State \textbf{return} S, P
\EndFunction

\end{algorithmic}
\end{algorithm}

Let us now describe the reward function used. Note that feature selection constitutes a multi-objective problem i.e. reducing the size of feature subset $|S|$ and at the same time increasing classification's performance $P$ (e.g. accuracy or F-score; performance metrics are discussed in Section~\ref{sec:metrics}). It should be emphasised that these two goals can be conflicting since reducing the number of features, valuable information can be lost.

Given a feature subset $S$, an upper boundary $b$ for the subset's size and its corresponding classification performance $P$, we propose the reward function shown in Equation~\ref{eq:rew_fun}: \begin{eqnarray}\label{eq:rew_fun}
r =
\begin{cases}
P & \text{if } |S| \leq b\\
P / cost & \text{if } |S| > b\\
\end{cases}
\end{eqnarray}
where $cost$ denotes the penalty for violating the upper boundary and defined as $cost = |S| / b$.

\begin{algorithm}[t]             
\caption{MARL Wrapper}   
\label{alg:fs_clean}                  
\begin{algorithmic}[1]           

\Function{FS\_MARL}{$\{T_1,...,T_k\}$}
	\Statex Input parameters: 
	\Statex $\{T_1,...,T_k\}$: training set's folds
	
	\State{initialise Q-values: $\forall a | Q(a) = -1$}
	\For{$episode = 1:num\_episodes$}
			\For{$agent = 1:num\_agents$}
				\If{$flag\_CLEAN == 1$}
					\State{execute greedy (online) action $a_i \in \{0,1\}$}
				\Else
					\State{execute $\epsilon$-greedy action $a_i \in \{0,1\}$}	
				\EndIf
			\EndFor
	
			\State observe joint action $a$
			\State get subset $S$ by considering only ``on'' actions
			\State get global reward $G(a) = Reward(S,\{T_1,...,T_k\})$\Comment Algorithm~\ref{alg:rew_fun}
	
			\For{$agent = 1:num\_agents$}
				\If{$flag\_CLEAN == 1$}
					\State{take $\epsilon$-greedy (offline) action $c_i \in \{0,1\}$}
					\State Calculate $C_i = G(a - a_i + c_i) - G(a)$
					\State Update $Q(c_i)\leftarrow Q(c_i)+\alpha[C_i - Q(c_i)]$
				\Else
					\State Update $Q(a_i)\leftarrow Q(a_i)+\alpha[G(a) - Q(a_i)]$
				\EndIf
			\EndFor
			
			\State{reduce $\alpha$ using $\alpha\_decay\_rate$}
			\State{reduce $\epsilon$ using $\epsilon\_decay\_rate$}
	\EndFor
	\State \textbf{return} S
\EndFunction

\end{algorithmic}
\end{algorithm}

\begin{algorithm}[t]             
\caption{Reward Function}   
\label{alg:rew_fun}                  
\begin{algorithmic}[1]           

\Function{Reward}{$S,\{T_1,...,T_k\}$}
	\State Input parameters: 
	\State - $S$: feature subset
	\State - $\{T_1,...,T_k\}$: training set's folds

	\For{j=1:k}
	     \State let the training set be T = $T_1 \bigcup T_{j-1} \bigcup T_{j+1} \bigcup T_k $ and validation set be $V = T_j$
	     \State learn hypothesis $h = L(T|_S)$
	     \State get performance of $h$ on $V|_S$
	\EndFor
	\State Let $P$ be the average performance
	
	\If{$|S| \leq b$}
		\State $r = P$
	\Else
		\State let $cost = |S| / b$
		\State $r = P / cost$	
	\EndIf
	\State \textbf{return} r
\EndFunction

\end{algorithmic}
\end{algorithm}

Since wrapper feature selection methods take into account the classification performance, we also describe in this section the ``k-fold cross-validation'' evaluation method we have used. Alternatives methods can be used, however, this is widely regarded as the standard one \cite{witten2011}. This method splits the training set into $k$ disjoint and stratified subsets, each of size $m/k$ (where $m$ is the number of training examples). Stratification is applied to ensure the same portion of positive examples is found in each split. Then, $k-1$ folds are used for training while the remaining fold is considered as the validation (also known as the cross-validation) set. This is repeated $k$ times so all folds are eventually considered as validation sets. Ideally, the learning process will come up with the feature subset that has the best average performance on all validation sets. The learnt feature subset will then be evaluated on the unseen test set.

In this novel formulation of feature selection the total number of learning agents is equal to the number of features. Rich datasets nowadays can include thousands of features. Therefore, this constitutes an enormous multiagent coordination problem and the scalability of the proposed approach is of vital importance. We refer to our approach that uses the global reward as MARL. To address the scalability challenge, we propose the use of CLEAN rewards as described in Section~\ref{sec:preliminaries_CLEAN}.

\begin{table*}[t]
	\begin{center}
	  \begin{tabular}{| c | c | c | c | c|}
	    \hline
	    \textbf{Dataset Name} & \textbf{\#Features} & \textbf{\#Examples} & \textbf{\#Positive Examples} \\ \hline\hline
	    Colon Cancer 	& 2000      & 62 		   & 40 (64.5\%)   \\\hline
	    Prostate Cancer & 2135  	    & 102 	   & 52 (51.0\%)   \\\hline
	    Leukemia 	    & 7129      & 72 		   & 25 (34.7\%)   \\\hline
	  \end{tabular}
	\end{center}

	\caption{Datasets used in our study}
	\label{tab:datasets}
\end{table*}

What has been described thus far is presented by Algorithms~\ref{alg:kFoldCV}-\ref{alg:rew_fun} in a top-down fashion. The feature selection process is initiated in Algorithm~\ref{alg:kFoldCV}. This algorithm calls the MARL wrapper method (Algorithm~\ref{alg:fs_clean}) which in turn makes use of the reward function in Algorithm~\ref{alg:rew_fun}.

\section{Experimental Setup}\label{sec:setup}

\subsection{Datasets}
For our experimental study we make use of three public microarray gene expression datasets. Microarray datasets contain a rich source of genetic data typically used for distinguishing between benign or malicious tissues or classifying different types of cancer. Microarray datasets are noisy, high-dimensional and constitute the ideal testbed for feature selection. The datasets used are for colon cancer \cite{alon1999}, prostate cancer \cite{glaab2012} and leukemia \cite{glaab2012}; more information is shown in Table~\ref{tab:datasets}.

We follow some typical data pre-processing steps prior applying any of the feature selection and classification algorithms. Since the range of numerical features can vary significantly, we apply mean normalisation to standardise these ranges. Specifically, for any feature $F_j$ let its feature space be $X|_{F_j}$. Let the mean and standard deviation of $X|_{F_j}$ to be denoted by $\mu$ and $\sigma$. Then each feature value in $\{x^{i}_{j}\}^{m}_{i=1}$ is normalised to $x^{i}_{j} = \frac{x^{i}_{j} - \mu}{\sigma}$. 

In addition, the mRMR feature selection method (described in Section~\ref{sec:related}) requires that numerical values are discretised.  As suggested by \cite{peng2005}, we use binning (3 bins) for data discretisation. Binning works as follows; the original feature values that fall in a given interval (i.e. a ``bin''), are replaced by a value representative of that interval.

\subsection{Compared Methods}
We compare a total of nine feature selection methods. Filter methods used are the univariate CFS (uCFS) and mRMR. Wrapper methods include genetic algorithms (GA) and the two proposed methods MARL and CLEAN. A description of uCFS, mRMR and GA was given in Section~\ref{sec:related}. The proposed MARL and CLEAN were described in Section~\ref{sec:MARL}. We further include in our comparison three hybrid methods, specifically, a combination of the wrappers GA, MARL and CLEAN with the filter uCFS. The last method is a baseline method where no feature selection is performed. We note that for the leukemia dataset, we first apply the uCFS filter as a pre-processing step to bring down the number of features to 2000.

Each method is run three times where the value of the upper boundary for the feature subset is $b = \{10,30,50\}$. The chosen values are based on studies which have showed that the approximate number of features to be selected in microarray datasets in order to obtain only genetic information with significant informative value is 30 \cite{glaab2012}.

\begin{figure*}[t]
\centering

\begin{subfigure}{.5\textwidth}
  \centering
  \includegraphics[scale=0.4]{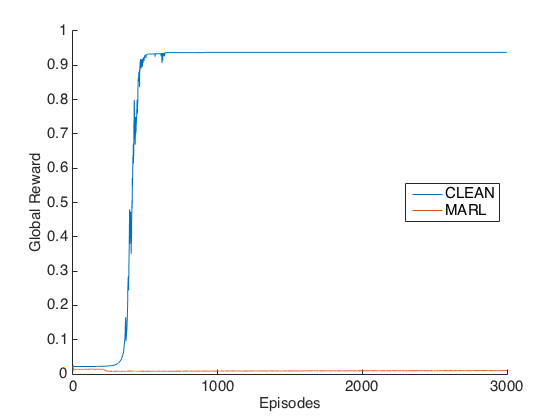}
  \caption{MARL \& CLEAN, $b=10$}
\end{subfigure}%
\begin{subfigure}{.5\textwidth}
  \centering
  \includegraphics[scale=0.4]{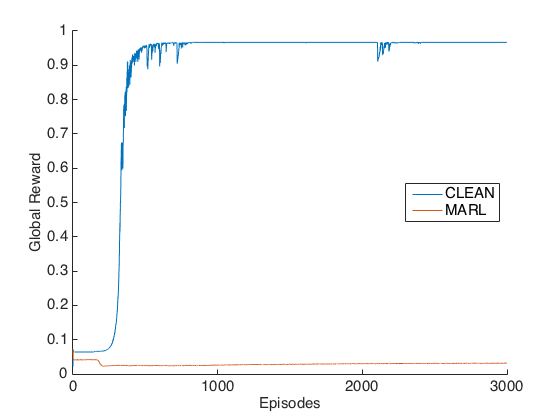}
  \caption{MARL \& CLEAN, $b=30$}
\end{subfigure}%

\begin{subfigure}{0.5\textwidth}
  \centering
  \includegraphics[scale=0.4]{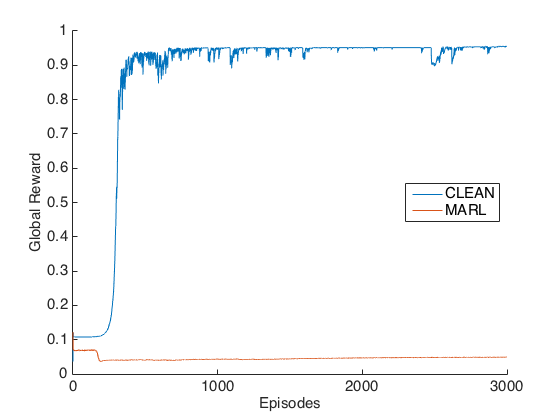}
  \caption{MARL \& CLEAN, $b=50$}
\end{subfigure}%
\begin{subfigure}{.5\textwidth}
  \centering
  \includegraphics[scale=0.4]{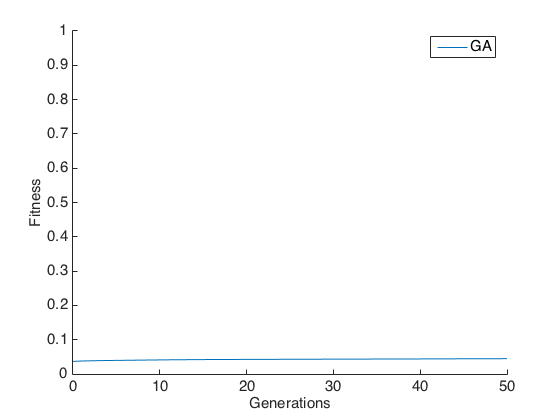}
  \caption{GA, $b=50$}
\end{subfigure}

\caption{Learning results (prostate cancer)}
\label{fig:learning}
\end{figure*}

Lastly, after performing parameter tuning we set the parameter values for MARL and CLEAN as follows: $\alpha = 0.2$, $\epsilon = 0.15$, $\alpha\_decay\_rate = 0.9995$, $\epsilon\_decay\_rate = 0.9995$. Also, the number of episodes for MARL is $num\_episodes = 5000$, while for CLEAN is $num\_episodes = 3000$.

The parameter values for GA are $population\_size = 50$, $num\_generations = 100$, $tournament\_size = 3$, two-point crossover with $prob\_crossover = 0.7$, $prob\_mutation = 1.0$ and $mutation\_rate = 1 / f$ where $f$ is the original number of features. The fitness function is the same as the reward function given in Equation~\ref{eq:rew_fun}.

\subsection{Evaluation Method and Metrics}\label{sec:metrics}
As discussed in Section~\ref{sec:MARL}, the evaluation method used is the ``stratified k-fold cross-validation'' \cite{witten2011}. We have set the split percentage for the final test set to $\lambda=0.2$ and number of folds to $k=10$. To ensure fairness, this process is repeated over ten times to obtain different splits.

The classification algorithm $L$ we have used is $k$-Nearest Neighbours \cite{witten2011}. Four popular evaluation metrics are used for classification tasks; these are, accuracy, precision, recall and F-score (also known as F1-score) \cite{sokolova2009}. Accuracy is the overall performance of the classification algorithm, precision is the ratio of true positives over the predicted positives, recall is the ratio of true positives over the actual positives and F-score is a trade-off between precision and recall.

Let TP and TN be the number of true positives and true negatives respectively, and let FP and FN be the number of false positives and false negatives respectively. The four metrics are defined as follows:
\begin{align}
	Accuracy & = \frac{TP + TN}{TP + FP + TN + FN}\\
	Precision (P) & = \frac{TP}{TP + FP}\\
	Recall (R) & = \frac{TP}{TP + FN}\\
	F-score & = \frac{2 \times P \times R}{P + R}
\end{align} 

For the final evaluation (Algorithm~\ref{alg:kFoldCV}/Line 6) we consider the values for all performance metrics. During learning the metric to be optimised is F-score (Algorithm~\ref{alg:rew_fun}/Line 10). Other metrics could have been used for learning, but our choice was based on the following fact. When accuracy was considered as the metric to be optimised, it resulted in poor results for F-score (and Precision and Recall). However, when F-score is optimised, it also results in good results for accuracy (although sometimes not as good as if accuracy was optimised directly). Therefore, since we are interested in obtaining a good all-round performance, F-score is used.

\subsection{Summary}
To sum up, three high-dimensional datasets (colon cancer, prostate cancer and leukemia) were considered, a baseline (without feature selection) and eight (uCFS, mRMR, GA, GA+uCFS, MARL, MARL+uCFS, CLEAN and\\CLEAN+uCFS) feature selection methods were compared, and each of these eight methods was run three times (with $b = \{10,30,50\}$). In total 75 experiments were conducted, and as mentioned, each was repeated over ten times. The experimental results are presented in the next sections.

\section{Learning Results}\label{sec:learning}
We present here the learning results for the three wrapper methods MARL, CLEAN and GA. Recall from Algorithm~\ref{alg:rew_fun}/Line 12 that the global reward corresponds to the average performance on the validation folds. However, if the subset size exceeds the desired boundary, a punishment is provided (Algorithm~\ref{alg:rew_fun}/Line 14).

We plot the global reward at each episode and values are averaged over ten repetitions. Due to space restrictions, Figure~\ref{fig:learning} depicts the results for the MARL and CLEAN ($b=\{10,30,50\}$) and GA ($b=50$) methods in the prostate cancer case; similar plots are obtained for the other datasets.

The first thing to notice is that the MARL and GA approaches behave similarly and extremely poorly. The reason behind the poor performance is because MARL and GA fail to scale-up as they always learn a feature subset of size significantly above the desired boundary $b$, and therefore big punishments occur. This will indeed be verified by the evaluation results on the unseen test set in the next section.

The superiority of the proposed CLEAN approach against the other wrapper methods is clearly apparent by the experimental outcome. The CLEAN approach can significantly scale-up to datasets of thousands of features. It is observed that as the boundary $b$ increases it takes more time to learn the desired feature subset, however, it always obtains an excellent performance and the learnt feature subset's size is always below or equal to $b$.

\section{Evaluation Results}\label{sec:eval}
We consider the feature subset learnt during the learning process and present the results on the unseen test set. Tables~\ref{tab:colon50} -~\ref{tab:leukemia10} show the results for colon cancer ($b=50$), prostate cancer ($b=30$) and leukemia ($b=10$) respectively; due to space restrictions the remaining six tables are not included. The figures are averaged over ten runs and the standard deviation is shown in the brackets. In the following sections, we will discuss in detail the results for each different objective (number of features and performance), but let us first describe the method used for analysing our results.

We have used the score-based analysis from \cite{bajic2000,soufan2015} to compare the different methods. The analysis works as follows. Consider, for instance, the column ``\#Features'' in Table~\ref{tab:leukemia10}. We give a score of 1 to the lowest (CLEAN+uCFS) method, the second lowest (GA+uCFS) receives 2 points, and so on. Therefore, the larger the number of features the higher the score; in this case, the lower the score the better.

For the performance objective there are a couple of minor differences. Firstly, according to \cite{bajic2000}, the performance measures should be statistically unrelated as much as possible. For this reason, we only consider accuracy, precision and recall but not the F-score. Secondly, the higher the performance the higher the score; therefore, in this case, the higher the score the better. The total scores and corresponding ranks for each objective are provided in Table~\ref{tab:ranks}.

\subsection{Number of Features}
Consider, for example, the Table~\ref{tab:leukemia10}. The first thing to notice is the huge feature subset size (about 800) for GA and MARL. It does of course constitute a significant reduction over the original size, but this is about 80 times above the requested subset size of $b=10$. This is in alignment with the learning results in Section~\ref{sec:learning}. The wrapper methods GA and MARL severely suffer from scalability issues.

The CLEAN method is the only wrapper that always keeps the subset size below the boundary $b$. The hybrid methods, as expected, reduce the subset size even further since they merge the benefits of both worlds. In Table~\ref{tab:ranks}, the proposed CLEAN+uCFS, MARL+uCFS and CLEAN methods are ranked first, third and fourth respectively.\\

\begin{table*}[p]
	\begin{center}
	  \begin{tabular}{| c || c || c | c | c | c |}
	  \hline
	  
\textbf{FS Method} &\textbf{\#Features} &\textbf{Accuracy} &\textbf{Precision} &\textbf{Recall} &\textbf{F-score} \\\hline\hline
Without FS &2000.0 (0.0) &71.5 (7.3) &73.0 (6.1) &87.5 (13.2) &78.9 (6.0) \\\hline\hline
uCFS &50.0 (0.0) &76.9 (6.3) &80.2 (6.1) &83.8 (8.4) &81.6 (5.1)  \\\hline
mRMR &50.0 (0.0) &78.5 (7.9) &82.1 (6.5) &83.8 (10.3) &82.6 (6.8) \\\hline\hline
GA &770.5 (23.9) &72.3 (11.6) &74.0 (9.6) &87.5 (13.2) &79.5 (8.5)  \\\hline
MARL &796.5 (83.4) &71.5 (6.3) &73.6 (6.8) &86.3 (12.4) &78.7 (5.2)  \\\hline
CLEAN &43.3 (2.9) &72.3 (13.2) &74.2 (12.0) &\textbf{87.5 (10.2)} &79.8 (9.2)  \\\hline\hline
GA + uCFS &23.9 (2.5) &76.9 (6.3) &79.4 (5.6) &85.0 (9.9) &81.8 (5.3)  \\\hline
MARL + uCFS &24.4 (4.4) &\textbf{80.0 (6.5)} &\textbf{82.4 (5.2)} &86.3 (9.2) &\textbf{84.0 (5.5)}  \\\hline
CLEAN + uCFS &\textbf{9.7 (3.7)} &76.2 (6.7) &79.4 (6.8) &83.8 (8.4) &81.2 (5.1)  \\\hline

	  \end{tabular}
	\end{center}

	\caption{Evaluation results for colon cancer ($b=50$)}
	\label{tab:colon50}
\end{table*}

\begin{table*}[p]
	\begin{center}
	  \begin{tabular}{| c || c || c | c | c | c |}
\hline

\textbf{FS Method} &\textbf{\#Features} &\textbf{Accuracy} &\textbf{Precision} &\textbf{Recall} &\textbf{F-score} \\\hline\hline
Without FS &2135.0 (0.0) &74.3 (6.8) &80.9 (10.0) &69.1 (13.7) &73.3 (8.0) \\\hline\hline
uCFS &30.0 (0.0) &87.6 (8.5) &\textbf{92.2 (7.0)} &83.6 (14.7) &87.1 (9.8) \\\hline
mRMR &30.0 (0.0) &87.6 (6.4) &92.1 (7.6) &84.5 (12.9) &87.4 (7.0) \\\hline\hline
GA &838.7 (15.3) &78.6 (7.2) &84.7 (11.3) &74.5 (11.2) &78.4 (7.2) \\\hline
MARL &886.5 (69.6) &78.6 (5.6) &82.7 (7.6) &76.4 (11.5) &78.6 (6.3) \\\hline
CLEAN &28.5 (0.7) &84.8 (5.9) &85.3 (5.5) &86.4 (11.5) &85.3 (6.4) \\\hline\hline
GA + uCFS &12.8 (2.9) &\textbf{89.0 (7.8)} &91.1 (8.0) &\textbf{88.2 (11.4)} &\textbf{89.2 (8.1)}  \\\hline
MARL + uCFS &16.1 (3.9) &88.6 (8.2) &92.2 (8.4) &86.4 (13.0) &88.5 (8.5)\\\hline
CLEAN + uCFS &\textbf{7.8 (2.3)} &86.7 (6.3) &90.0 (6.6) &84.5 (12.2) &86.6 (7.0)\\\hline

	  \end{tabular}
	\end{center}
	\caption{Evaluation results for prostate cancer ($b=30$)}
	\label{tab:prostate30}
\end{table*}

\begin{table*}[p]
	\begin{center}
	  \begin{tabular}{| c || c || c | c | c | c |}
	  \hline
	  
\textbf{FS Method} &\textbf{\#Features} &\textbf{Accuracy} &\textbf{Precision} &\textbf{Recall} &\textbf{F-score}  \\\hline\hline
Without FS &7129.0 (0.0) &84.0 (5.6) &94.2 (12.5) &56.0 (12.6) &69.5 (11.6)  \\\hline\hline
uCFS &10.0 (0.0) &88.0 (8.2) &86.3 (12.1) &76.0 (18.4) &80.1 (13.9)  \\\hline
mRMR &10.0 (0.0) &90.0 (8.5) &85.2 (10.7) &\textbf{84.0 (18.4)} &84.2 (14.1)  \\\hline\hline
GA &806.5 (30.0) &90.7 (7.2) &95.5 (9.6) &76.0 (18.4) &83.4 (13.8) \\\hline
MARL &770.9 (78.3) &\textbf{93.3 (7.0)} &\textbf{100.0 (0.0)} &80.0 (21.1) &\textbf{87.4 (14.4)} \\\hline
CLEAN &9.3 (0.5) &84.0 (5.6) &75.5 (11.8) &82.0 (11.4) &77.6 (6.6) \\\hline\hline
GA + uCFS &5.1 (1.8) &88.7 (8.3) &87.8 (14.2) &78.0 (17.5) &81.6 (13.3) \\\hline
MARL + uCFS &5.7 (1.3) &86.7 (8.9) &85.3 (14.1) &74.0 (21.2) &77.6 (15.5)  \\\hline
CLEAN + uCFS &\textbf{3.9 (0.9)} &85.3 (6.9) &82.1 (13.7) &74.0 (16.5) &76.7 (11.1) \\\hline

	  \end{tabular}
	\end{center}

	\caption{Evaluation results for leukemia ($b=10$)}
	\label{tab:leukemia10}
\end{table*}

\begin{table*}[p]
	\begin{center}
	  \begin{tabular}{| c || c || c |}
	    \hline
	    \textbf{FS Method} & \textbf{\#Features Score} & \textbf{Performance Score} \\\hline\hline
     
    	    Without FS   &  81 (9th)  & 71  (9th)  \\\hline\hline
    	    uCFS         &  45 (5th)  & 157 (4th)  \\\hline
    	    mRMR         &  45 (5th)  & 160 (3rd)  \\\hline\hline
    	    GA 			 &  66 (7th)  & 110 (7th)  \\\hline
    	    MARL 		 &  69 (8th)  & 107 (8th)  \\\hline
    	    CLEAN 	     &  36 (4th)  & 113 (6th)  \\\hline\hline
    	    GA + uCFS    &  19 (2nd)  & 164 (2nd)  \\\hline
     	MARL + uCFS  &  26 (3rd)  & \textbf{169 (1st)}  \\\hline
	    CLEAN + uCFS &  \textbf{9 (1st)}   & 133 (5th)  \\\hline
   
	  \end{tabular}
	\end{center}

	\caption{Overall evaluation scores and rankings}
	\label{tab:ranks}
\end{table*}

\subsection{Performance}
Consider, for example, the Table~\ref{tab:prostate30}. The first thing to notice is how important feature selection is. The baseline approach with 2135 features has a larger amount of available information, but its performance is extremely poor. This is observed in all tables as summarised by Table~\ref{tab:ranks} where the baseline method is ranked last.

A similar behaviour is observed for the GA and MARL wrappers. Although they use a significantly larger amount of information (subset size of about 800) their ranking is 7th and 8th respectively, while their hybrid variants have the best overall performance. The proposed MARL+uCFS is ranked first, while GA+uCFS is ranked second. The filters follow in the 3rd and 4th positions. As expected, the univariate filter (uCFS) performs worse than the multivariate filter (mRMR).

The proposed CLEAN method outperforms the baseline and the other wrapper (GA,MARL) methods. However, CLEAN and its hybrid variant only rank 5th and 6th overall. This is a surprising outcome based on the learning results from Figure~\ref{fig:learning} where the CLEAN method always obtains a global reward of over 0.9. We believe this result occurs because of over-fitting to the validation sets, despite the fact that the stratified 10-fold cross-validation method was used. We believe over-fitting is attributed to the fact that the number of examples in our datasets is very limited (Table~\ref{tab:datasets}). We plan in the future to investigate datasets with a larger number of examples.

\section{Conclusion}\label{sec:conclusion}
We have formulated feature selection as a multiagent coordination problem and proposed a novel wrapper method using multiagent reinforcement learning. The central idea of the proposed approach is to ``assign'' a reinforcement learning agent to each feature. Each agent controls a single feature and learns whether it will be included in or excluded from the final feature subset; we referred to this approach as MARL. Furthermore, we have proposed the incorporation of CLEAN rewards, a form of reward shaping, that removes the exploratory noise created by other learning agents and has been shown to significantly scale-up.

Our experimental setup involved the use of three noisy and high-dimensional (thousands of features) microarray datasets, namely, colon cancer, prostate cancer and leukemia. CLEAN is the only wrapper that can learn a feature subset with a size below the desired boundary; GA and MARL severely suffer from scalability issues. In addition, CLEAN outperforms the baseline, GA and MARL methods despite the fact that these three methods take into consideration more features and therefore more information is available to them.

Due to over-fitting, CLEAN is outperformed by the filter methods. It is believed that occurred because of the limited number of examples in the microarray datasets. We plan in the future to investigate a variety of datasets. Furthemore, despite the fact that MARL is inferior to CLEAN, its hybrid variant (MARL+uCFS) achieves the best overall performance. Future work will further investigate streaming data where we believe the true power of reinforcement learning will be revealed.

\section*{Acknowledgements}
We would like to thank Mitchell Colby and Sepideh Kharaghani for help on CLEAN rewards. We would also like to thank Tamas Jambor, Charlie Evans and Vigginesh Srinivasan for the fruitful discussions we have had.

%
%
\bibliographystyle{abbrv}
\bibliography{bibfile}  
%
\end{document}